\documentclass[a4paper]{article}
\usepackage{interspeech2013,amssymb,amsmath,epsfig}
\usepackage{algorithm,algorithmic,cite,amsmath,amssymb,epsfig,enumerate,amsfonts,url,multirow,array,tabularx,booktabs,stfloats,flushend}
\usepackage[T1]{fontenc}

\def\c{{\mathbf c}}

\def\W{{\mathbf W}}

\floatname{algorithm}{Scheme}

\ninept
\def\reg{{\rm\ooalign{\hfil
     \raise.07ex\hbox{\scriptsize R}\hfil\crcr\mathhexbox20D}}}

\title{Transfer Learning for Voice Activity Detection: A Denoising Deep Neural Network Perspective}

\makeatletter
\def\name#1{\gdef\@name{#1\\}}
\makeatother
\name{{\em Xiao-Lei Zhang, Ji Wu}}

\address{Multimedia Signal and Intelligent Information
Processing Laboratory, \\
Tsinghua National Laboratory for Information Science and Technology,\\
Department of Electronic Engineering,
Tsinghua University, Beijing, China.\\
{\small \tt huoshan6@126.com, wu\_ji@tsinghua.edu.cn}
\thanks{This work is supported in part by the China Postdoctoral Science Foundation funded project under Grant 2012M520278, and in part by the National Natural Science Funds of China under Grant 61170197.}
}

\begin{document}
\maketitle
\begin{abstract}
Mismatching problem between the source and target noisy corpora severely hinder the practical use of the machine-learning-based voice activity detection (VAD).
 In this paper, we try to address this problem in the transfer learning prospective. Transfer learning tries to find a common learning machine or a common feature subspace that is shared by both the source corpus and the target corpus. The denoising deep neural network is used as the learning machine. Three transfer techniques, which aim to learn common feature representations, are used for analysis. Experimental results demonstrate the effectiveness of the transfer learning schemes on the mismatch problem.
\end{abstract}

\section{Introduction}
\label{sec:intro}
Voice activity detectors (VADs) aim to discover speech from its background noises. They are important frontends of modern speech recognition systems \cite{yu2010deep,dahl2012context,hinton2012deep} and speech signal processing systems \cite{hanaclassification}. Recently, the machine-learning-based VADs \cite{yu2010discriminative,wu2011mmc,wu2011efficient,zhang2012linearithmic,suh2012multiple} have received much attention in that they not only can be integrated to the speech recognition systems naturally but also can fuse the advantages of multiple features \cite{wang2005time,wang2006computational,han2011towards,wang2012exploring,hu2010unsupervised,wang2013towards} much better than traditional VADs.
However, the machine-learning-based VAD is still far from its practical use. One significant problem is that we are not sure whether the VAD model trained in a given source corpus is still powerful in a target corpus which might have a different distribution with the source corpus.

In this paper, we try to deal with the aforementioned problem by a novel learning method -- transfer learning. Generally, transfer learning tries to make the model trained with one or multiple source tasks generalizes well on different but related target tasks, so that the performance gap between the source tasks and the target tasks can be lowered. See \cite{pan2010survey} for an excellent survey on transfer learning. In respect of \textit{different hypothesis} on whether the source data or target data is manually labeled, the transfer learning technologies can be categorized into four groups \cite{pan2010survey}.  This paper focuses on the \textit{domain adaptation} techniques, where the source data is manually labeled and the target data is unlabeled which is a practical scene that the machine-learning-based VAD will meet.
In respect of \textit{what to transfer}, the transfer learning methods can be categorized to three groups -- instance transfer, feature transfer, and parameter transfer (i.e. model transfer) \cite{pan2010survey,sha2012domain}. This paper focuses on the feature transfer techniques. Generally, feature transfer tries to learn a low-dimensional feature representation that is shared by both the source data and the target data, so that the classifier trained on the source data with the learned subspace can generalize well on the target data.
The main contributions of this paper are summarized as follows:
\begin{enumerate}
  \item \textbf{Towards the mismatching problem of the machine-learning-based VAD.} We have conducted an extensive experiment from the domain adaptation perspective for the mismatching problem. The recently proposed denoising deep neural network (DDNN) \cite{zhang2013denoising} is used as the learning machine. Empirical results show that the transfer learning schemes are more powerful than several state-of-the-art VADs when the source and target corpora are relatively similar. The results also demonstrate the promising future of the practical use of the machine-learning-based VADs.
  \item \textbf{A useful empirical comparison of three feature-based domain adaptation schemes.} We have proposed three domain adaptations for the DDNN-based VAD. Empirical results show that we can pre-train the deep neural networks in an unsupervised manner either with the source data only or with both the source data and the target data together, but the data for all layers' pre-training have to be the same without interference, which manifested the powerfulness of the pre-training scheme proposed by Hinton \textit{et al.} \cite{hinton2006reducing}.
\end{enumerate}
We have to note that the main purpose of this paper is to discuss the effectiveness of the transfer learning for the mismatching problem between the source data and the target data but not to propose a specific VAD algorithm. To make the machine-learning-based VAD work well in practice, a lot of efforts are still needed. As an example, the DDNN-based VAD needs the clean speech signals of its noisy speech corpus in the unsupervised pre-training stage, which is an ideal situation.

The remainder of the paper is organized as follows. In Section \ref{sec:svm}, we first review the recently proposed DDNN and then present three feature-based domain adaptation schemes for the DDNN-based VAD. In Section \ref{ssec:subsubhead}, we present the related work. In Section \ref{sec:analysis}, we conduct an extensive experimental comparison. In Section \ref{sec:conclusion}, we conclude this paper with some future work.

\section{Domain Adaptation for VAD}\label{sec:svm}
In this section, we first review the denoising deep neural network (DDNN), and then propose three feature-based domain adaptation schemes for the DDNN-based VAD.

\subsection{Review: Denoising Deep Neural Network for VAD}

DDNN \cite{zhang2013denoising} is a deep neural network. It was motivated from the stacked denoising autoencoder \cite{vincent2008extracting,vincent2010stacked}. Compared to the deep-belief-network-based VAD \cite{zhang2013deep}, it has achieved a success on the performance of the deep layers over shallower layers. The key idea of DDNN is to first minimize the \textit{reconstruction cross-entropy loss} between the noisy speech signal and its corresponding clean speech signal in an unsupervised greedy layer-wise pre-training way, and then fine-tune the entire deep neural network by minimizing the cross-entropy loss between the noisy speech signal and its manual labels for the minimum classification error. One special point of DDNN is that, in the pre-training phase, DDNN needs to train an accompanying deep neural network, i.e. a deep network that tries to reconstruct the clean speech signal from the clean speech signal. This is mainly to supply the noisy speech its optimization objective in each layer.

From the aforementioned, we can see that one weakness of DDNN is that the noisy speech signal needs its corresponding clean speech signal in the pre-training phase. Because this paper focuses on the effectiveness of the transfer learning, this weakness does not hinder the main contributions of the paper.

\begin{algorithm}[t]
    \caption{ A successful scheme.}
    \begin{algorithmic}[1]\label{alg:1}
\STATE Pre-train all layers of DDNN with only the target corpus $\mathcal{X}^{(t)}$.
\STATE Fine-tune the pre-trained DDNN with only the labeled source corpus, i.e. $\mathcal{X}^{(s)}\times\mathcal{Y}^{(s)}$.
\end{algorithmic}
\end{algorithm}

    \begin{algorithm}[t]
    \caption{A successful scheme.}
    \begin{algorithmic}[1]\label{alg:2}
\STATE Take the source corpus $\mathcal{X}^{(s)}$ and the target corpus $\mathcal{X}^{(t)}$ together as a large corpus.
 \STATE Pre-train all layers of DDNN with the large corpus.
\STATE Fine-tune the pre-trained DDNN with only the labeled source corpus, i.e. $\mathcal{X}^{(s)}\times\mathcal{Y}^{(s)}$.
\end{algorithmic}
\end{algorithm}

\begin{algorithm}[t]
    \caption{A failed scheme.}
    \begin{algorithmic}[1]\label{alg:3}
\REQUIRE  The desired depth of DDNN, denoted as $L$, (i.e. the hidden-layer number).
\STATE \textbf{Source DDNN pre-training: }Pre-train the source DDNN with a depth of $L-1$ with only $\mathcal{X}^{(s)}$. The pre-trained source DDNN is denoted as $\left\{\W_l^{(s)}\right\}_{l=1}^{L-1}$. /*Note: this model needs to be trained only once, and used repeatedly for different target corpus.*/
\STATE \textbf{Target DDNN pre-training: }Pre-train the source DDNN with a depth of $L-1$ with only $\mathcal{X}^{(t)}$. The pre-trained target DDNN is denoted as $\left\{\W_l^{(t)}\right\}_{l=1}^{L-1}$.
 \STATE \textbf{Hybrid pre-training of the top layer}: 
 Group the output features of the source DDNN and target DDNN together to a large set, and pre-train the $L$-th layer of DDNN with the large set. The pre-trained model is denoted as $\W_L^{(t)}$.
 \IF{Scheme 3$^{(t)}$}
\STATE Fine-tune the pre-trained $\left\{\left\{\W_l^{(t)}\right\}_{l=1}^{L-1},  \W_L^{(t)}\right\}$ with only the labeled source corpus, i.e. $\mathcal{X}^{(s)}\times\mathcal{Y}^{(s)}$.
\ELSIF{Scheme 3$^{(s)}$}
\STATE Fine-tune the pre-trained $\left\{\left\{\W_l^{(s)}\right\}_{l=1}^{L-1},  \W_L^{(t)}\right\}$ with only the labeled source corpus, i.e. $\mathcal{X}^{(s)}\times\mathcal{Y}^{(s)}$.
\ENDIF
\STATE Output the fine-tuned network as the learned DDNN.
\end{algorithmic}
\end{algorithm}

\subsection{Preliminary of the Feature-Based Domain Adaptation}
Suppose we have a labeled source corpus $\mathcal{X}^{(s)}\times\mathcal{Y}^{(s)} $, and an unlabeled target corpus $\mathcal{X}^{(t)} $, where $\mathcal{X}$ denotes the acoustic feature corpus and $\mathcal{Y}$ denotes the set of the manual labels. The corpora $\mathcal{X}^{(s)}$ and $\mathcal{X}^{(t)}$ might be sampled from different noise scenarios.
The feature-based domain adaptation scheme aims to find a mapping function $\phi(\cdot)$ such that the distribution difference between $\phi\left(\mathcal{X}^{(s)}\right)$ and $\phi\left(\mathcal{X}^{(t)}\right)$ is minimized. Therefore, if we minimize the classification error on the source corpus with the new feature representation, i.e. $\phi\left(\mathcal{X}^{(s)}\right)$, we can also expect to minimize the classification error on the target corpus.

\subsection{Domain Adaptation Via Deep Feature Extraction}

For the DDNN-based VAD, a number of training schemes for $\phi(\cdot)$ can be developed. The core idea of the development is to first pre-train DDNN in different unsupervised ways and fine-tune DDNN with the labeled source corpus, i.e. $\mathcal{X}^{(s)}\times\mathcal{Y}^{(s)} $, for the minimum classification error.
 
 In this paper, we present three unsupervised pre-training schemes, which are described in Schemes \ref{alg:1}, \ref{alg:2}, and \ref{alg:3} respectively. 
The effectiveness and efficiency of the three schemes are analyzed qualitatively as follows:

 Scheme \ref{alg:1} only uses the unlabeled target data $\mathcal{X}^{(t)}$ to initialize DDNN. Because it uses only $\mathcal{X}^{(t)}$ for pre-training, it is supposed to be a relatively poor initialization scheme but computationally efficient. Moreover, when $\mathcal{X}^{(t)}$ is small, the initial point of DDNN might be biased and still suffer from overfitting, hence, the network might not be trained well.

Scheme \ref{alg:2} uses both $\mathcal{X}^{(s)}$ and $\mathcal{X}^{(t)}$ for initializing DDNN, which can learn a good feature representation shared by $\mathcal{X}^{(s)}$ and $\mathcal{X}^{(t)}$. Particularly, when $\mathcal{X}^{(t)}$ is rare, $\mathcal{X}^{(s)}$ can play a sufficient supplementary role to $\mathcal{X}^{(t)}$. Hence, the network is desired to perform gently well on the target test data. However, whenever we meet a new target task, we have to conduct a heavy computation load by training $\mathcal{X}^{(s)}$ and $\mathcal{X}^{(t)}$ jointly in the pre-training phase.

 Scheme \ref{alg:3} is designed to be a compromise between Scheme \ref{alg:1} and Scheme \ref{alg:2}. Specifically, because we take the supplementary effect of $\mathcal{X}^{(s)}$ merely into the highest layer of DDNN which is a layer that directly influences the performance of DDNN, we might not only transfer the source knowledge to the target domain but also can save a lot of training time, since that 1) the most computationally expensive part of DDNN is the pre-training of the source DDNN which can be trained once for all; 2) the top layer of DDNN usually has much less hidden units (i.e. much less training time) than the bottom modules. Scheme \ref{alg:3} contains two sub-schemes, which is denoted as Scheme 3$^{(t)}$ and Scheme 3$^{(s)}$ respectively.
 
 Before the experimental section, we emphasize that 1) Schemes \ref{alg:1} and \ref{alg:2} are successful ones because the data for all layers' pre-training are the same, 2) Scheme \ref{alg:3} fails in providing a good initial point for DDNN, because the data for pre-training all layers is not consistent. The main purpose that we want to share the failed scheme is to tell the critical readers that it provides a compromise thinking between the computationally light Schemes \ref{alg:1} and the computationally heavy Scheme \ref{alg:2}, and we might find a successful compromise scheme that is both as effective as Scheme \ref{alg:2} and as efficient as Scheme \ref{alg:1} in the future.

Note that we can use multiple source corpora and multiple target corpora together to train the model freely. But in this paper, we only discuss the empirical performance with one source corpus and one target corpus, leaving the multiple source domain adaptation problem to a future discussion.

\section{Related Work}\label{ssec:subsubhead}
In respect of transfer learning and deep learning, there has been some similar work with the proposed schemes. For example, in \cite{glorot2011domain}, Glorot {\textit{et al.}} adopted a domain adaptation scheme that is exactly the same as Scheme 2 of this paper for the sentiment classification problem. In \cite{collobert2008unified}, Collobert and Weston proposed a joint training scheme for the multitask learning problem of natural language processing, whose key idea is similar with Scheme 3. The architecture of \cite{collobert2008unified} is also successfully applied to machine translation \cite{deselaers2009deep}. However, Scheme 3 is different from \cite{collobert2008unified,deselaers2009deep} in that our Scheme 3 pre-train the top hidden-layer of DDNN with set $\left\{\mathcal{X}^{(s)},\mathcal{X}^{(t)} \right\}$, while the architecture of \cite{collobert2008unified,deselaers2009deep} try to learn a subspace of word mapping in the top-hidden layer with a strong constraint that one word in the lookup table of $\mathcal{X}^{(s)}$ should have a matching word in that of $\mathcal{X}^{(t)} $.

In respect of the VAD study, the distribution difference between different noise scenarios has been mentioned in traditional VADs. For example, in \cite{chang2006voice}, Chang \textit{et al.} used different statistical models for modeling the speech and noise distributions in different noise scenarios. Another related topic with domain adaptation is the online learning methods \cite{ying2011voice}, they update the model parameters according to the historical domain information of the speech signals. Traditional statistical-model-based VADs \cite{chang2006voice} can also be regarded as unsupervised online learning methods. But to our knowledge, how to combine multiple features effectively is still an open problem in the online learning methods. On the other side, although the domain-adaptation-based VAD works in batch mode, it can combine multiple features effectively and yield a high accuracy without a requirement of heavy manual labeling.

\section{Experiments}\label{sec:analysis}

Seven noisy test corpora of AURORA2 \cite{pearce2000aurora} is used for performance analysis. The signal-to-noise ratio level of the audio signals is set to $5$ dB. Each test corpus of AURORA2 contains 1001 utterances, which are split randomly into three groups for training, developing and test respectively. Each training set and development set consist of 300 utterances respectively. Each test set consists of 401 utterances.

  The sampling rate is 8kHz. We set the frame length to 25ms long with a frame-shift of 10ms.
We extract 10 acoustic features from each observation. The detailed information of the features are listed in Table \ref{tab:feature}. All features are normalized into the range of $[0,1]$ in dimension.

 \begin{table} [t]
\caption{\label{tab:feature} {Features and their attributes. The subscript of each feature is the window length of the feature \cite{ramirez2005statistical}.
}}
\centerline{
\scalebox{0.6}{
\begin{tabular}{|l|c|c||l|c|c|}
 \hline
\textbf{ID} & \textbf{Feature} & \textbf{Dimension} & \textbf{ID}&	 \textbf{Feature}& \textbf{Dimension}\\
 \hline
1 & Pitch &  1   &		7 & MFCC$_{16}$&20	\\
\hline
2 & DFT & 16 &8&LPC	&12	\\
 \hline
3& DFT$_8$ &  16  & 9&	RASTA-PLP	&17	\\
 \hline
4& DFT$_{16}$&16	&10	&AMS	&	135\\
 \hline
5 & MFCC&  20  &	& \textbf{Total} &273	\\
 \hline
6 & MFCC$_{8}$&  20   &	& &\\
 \hline
\end{tabular}}}
\end{table}

To simulate the real-world domain adaptation task, we take the training sets of the \textsf{Street} and \textsf{Subway} noise scenarios as two source corpora. For each source corpus, we form 6 domain adaptation tasks by randomly extracting a 30-second audio segment from the training set of each noise type of AURORA2 except that of the source corpus.
For each domain adaptation task, the development set of the source corpus is used for model selection. We run each domain adaptation task 5 times and report the average accuracies.

\begin{table*} [thb]
\caption{\label{tab:time} {Transfer accuracy comparison (in percentage) with the \textsf{street} noise corpus (identification = 4) as the source data. ``LB'' is short for the lower bound, ``S1'' is short for Scheme 1, ``S2'' is short for Scheme 2, ``S3$^{(t)}$'' is short for Scheme 3$^{(t)}$, ``S3$^{(s)}$'' is short for Scheme 3$^{(s)}$, and ``UB'' is short for upper bound. ``\# layers'' means that the depth of the DDNN is ``\#''. Because the experimental environment settings are exactly as \cite{zhang2013deep} did, we just copy the results of the referenced VADs from \cite{zhang2013deep}. Due to the length limit, we only report the best performance of the referenced VADs and its corresponding VAD algorithm. The referenced methods that are marked with ``*'' means that they are machine-learning-based VADs that are trained and tested in the matching environments.}}
\centerline{
\scalebox{0.62}{
\begin{tabular}{|l|l||l||c|c|c|c||c|c|c|c|c|c||c|c|c|c|c|c|}
\hline
\multirow{2}{*}{\textbf{ID}}&\multirow{2}{*}{\textbf{Noise Type}}&\multirow{2}{*}{Referenced}& \multicolumn{4}{c|}{{1 layer}} & \multicolumn{6}{c|}{{2 layers}} & \multicolumn{6}{c|}{\textsf{3 layers}}\\
\cline{4-19}
 && & LB& S1 & S2 & UB    & LB & S1 & S2 & S3$^{(t)}$ & S3$^{(s)}$ & UB    & LB& S1 & S2 & S3$^{(t)}$ & S3$^{(s)}$ & UB    \\
 \hline
1 &\textsf{Babble} &75.51 (Ramirez05)&  74.95 &\textbf{77.15}  & 76.44 & \textit{78.61}& 74.09  & 75.67 & 76.59 & 75.73 &73.17  & \textit{78.85}& 72.72 & 75.53 & 75.92 & 73.74 & 72.84  & \textit{79.14} \\
 \hline
2& \textsf{Car} &79.25 (G.729B) &  81.89& 82.91 & \textbf{83.51} &  \textit{86.77}& 82.05  & 82.08 & 83.17 & 82.19 &  81.73& \textit{86.96} & 81.49  & 81.81 & 82.92 & 81.11 &  82.20  &\textit{87.09}\\
 \hline
3 &\textsf{Restaurant}& 69.59 (Ramirez05) & 74.44  & 75.14  & \textbf{75.74} & \textit{83.47}& 73.84   &75.34  & 75.61 & 74.53 & 74.17 &\textit{83.90}&73.25  & 75.59 & 75.19 & \textbf{75.76} &  73.31 &\textit{83.78 } \\
 \hline
5 & \textsf{Airport} &72.45 (Shin)*& 77.35 & \textbf{77.92}  & 77.56 &  \textit{82.34}& 77.12  & 77.82 &  \textbf{77.88} & 77.51 & 77.32 & \textit{82.45} &76.73   & 77.34 & \textbf{77.86}& 77.18 &  76.96  &\textit{82.30}\\
 \hline
6 &\textsf{Train} & 75.26 (G.729B)& 80.51& 81.69   & 81.22 & \textit{83.88}& 80.47  &  80.64&  \textbf{82.27} &  81.42& 80.88 & \textit{84.21} &79.70  & 80.76 & 81.89&  80.50  & 79.90 & \textit{84.25}\\
 \hline
7 & \textsf{Subway} & 73.16 (Ramirez05)& 68.19 & 68.19   & 69.87 &\textit{85.84}& 68.35  & 72.69 & 76.28 & 68.22 & 68.17 & \textit{85.62} &68.44  & 74.49 & \textbf{76.42}&  70.70  & 68.26 &\textit{ 85.73}\\
 \hline
\end{tabular}}
}
\end{table*}

\begin{table*} [thb]
\caption{\label{tab:time2} {Transfer accuracy comparison (in percentage) with the \textsf{subway} noise corpus (identification = 7) as the source data.}}
\centerline{
\scalebox{0.62}{
\begin{tabular}{|l|l||l||c|c|c|c||c|c|c|c|c|c||c|c|c|c|c|c|}
\hline
\multirow{2}{*}{\textbf{ID}}&\multirow{2}{*}{\textbf{Noise Type}}&\multirow{2}{*}{Referenced}& \multicolumn{4}{c|}{{1 layer}} & \multicolumn{6}{c|}{{2 layers}} & \multicolumn{6}{c|}{\textsf{3 layers}}\\
\cline{4-19}
 && & LB& S1 & S2 & UB    & LB & S1 & S2 & S3$^{(t)}$ & S3$^{(s)}$ & UB    & LB& S1 & S2 & S3$^{(t)}$ & S3$^{(s)}$ & UB    \\
 \hline
1 &\textsf{Babble}  &\textbf{75.51} (Ramirez05)&  54.58 & 54.60 & 62.05 & \textit{78.61}& 54.58  & 54.59 & 67.59 & 54.58 & 54.58 & \textit{78.85}& 54.58 &54.58 & {68.11} & 54.59 &  54.58 & \textit{79.14} \\
 \hline
2& \textsf{Car} &\textbf{79.25} (G.729B) &  55.80& 55.80 & 68.24 &  \textit{86.77}& 59.54  & 58.05 & 69.19 & 63.09 &64.11  & \textit{86.96} & 61.33  & 57.96 &{70.05} & 56.52 & 58.65   &\textit{87.09}\\
 \hline
\end{tabular}}
}
\end{table*}

The parameters are set as follows. Up to three hidden layers are adopted with the numbers of the hidden units set to $[54,7,7]$ respectively. The learning rate of the unsupervised pre-training is set to 0.004. The maximum epoch of the unsupervised pre-training is set to 200. The learning rate of the supervised fune-tuning is set to 0.005. The maximum epoch of the supervised fune-tuning is set to 130. The batch mode training is adopted. Each batch contains 512 observations.
Note that the parameters are selected empirically for a compromise between the training time and the accuracy. The accuracy might be further improved by tuning the parameters.

To evaluate the effectiveness of the proposed domain adaptation schemes, we give the empirical lower bound and upper bound that the schemes might achieve. The lower bound, denoted as ``LB'', is obtained by training DDNN with only the source corpus and testing it on various target environments. If the performance of the proposed domain adaptation schemes is worse than LB, it means that the schemes fail. The performance upper bound, denoted as ``UB'', is obtained by training DDNN with the training set of the target corpus and testing it on the test set of the same target environments. If the performance of the proposed domain adaptation schemes is better than the UB, it means that the schemes achieve unbelievably amazing successes.
We also compare with the G.729B VAD \cite{benyassine1997itu}, ETSI advanced frontend via Wiener filter  \cite{processing202transmission}, ETSI advanced frontend via frame dropping \cite{processing202transmission}, Sohn VAD \cite{sohn1999statistical}, Ramirez05 VAD \cite{ramirez2005statistical}, Ramirez07 VAD \cite{ramirez2007improved}, Yu VAD \cite{yu2010discriminative}, Shin VAD \cite{shin2010voice}, and Ying VAD \cite{ying2011voice}. The experimental settings are exactly as \cite{zhang2013deep} did.

\subsection{Experimental Results}
\label{sec:perform}

\begin{figure}[!t]
\centerline{\epsfig{figure=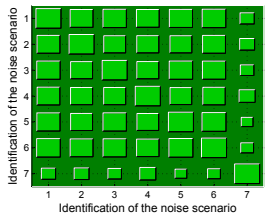,width=40mm}}
\caption{{Hinton diagram of the feature distributions in different noise scenarios. Each grid of the Hinton diagram measures the distribution similarity of the features in the relevant two scenarios. The bigger the grid is, the more similar the two distributions are. The similarity is calculated as $\exp\left(-\|\c^{(s)}-\c^{(t)} \|^2/2 \right)$ \cite{liu2009semisupervised} with $\c$ as the feature centroid.}}
\label{fig:fig4}
\end{figure}

First, we give the Hinton diagram of the feature distributions in different noise scenarios in Fig. \ref{fig:fig4}. From the figure, we can see that most feature distributions are relatively similar with each other except the \textsf{subway} noise scenario, which means the transfer learning schemes might be useful.

Table \ref{tab:time} lists the transfer accuracies with the \textsf{street} noise as the source corpus. From the figure, we can see that in all layers, Scheme 2 is the most powerful one, followed by Scheme 1. Both Scheme 2 and Scheme 1 achieve higher accuracies than LB, which means the \textit{positive transfer} \cite{pan2010survey} phenomenon is observed. However, both Scheme 3$^{(t)}$ and Scheme 3$^{(s)}$ are not only worse than Schemes 1 and 2, but also sometimes slightly worse than LB, which means the \textit{negative transfer} \cite{pan2010survey} is observed. This phenomenon is rather important. It manifested empirically that using the greedy layer-wise pre-training to initialize the deep network is valuable. If we interrupt the initial point of some layer by noises, or if the data for pre-training are inconsistent in all layers, the performance drops dramatically.

 Table \ref{tab:time2} lists the transfer accuracies with the \textsf{subway} noise as the source corpus. Due to the length limit, we only show the results of two target noise corpora. The experimental phenomena in other noise scenarios are similar with the two. From the table, we can see that due to the significant difference between the \textsf{subway} noise and the target corpus, the accuracies of all schemes drop significantly from UB. However, we can also observe that the accuracies yielded from Scheme 2 are still significantly better than LB and are upgraded layer by layer, which means that the positive transfer is observed too. 
 
 As a conclusion, 1) the proposed schemes are effective in dealing with the mismatching problem between the source data and the target data; 2) Schemes 1 and 2 are both effective transfer learning schemes; 3) Initialization via unsupervised greedy layer-wise pre-training is valuable.

\begin{table} [t]
\caption{\label{tab:time3} {Pre-training time (in seconds) comparison.}}
\centerline{
\scalebox{0.62}{
\begin{tabular}{|l|c|c|c|c|c|c|}
 \hline
 & \multicolumn{2}{c|}{1 layer}& \multicolumn{2}{c|}{2 layers} & \multicolumn{2}{c|}{3 layers}\\
 \hline
S1 & \multicolumn{2}{c|}{ 570.48} &  \multicolumn{2}{c|}{713.21 } & \multicolumn{2}{c|}{774.87 } \\
\hline
S2 & \multicolumn{2}{c|}{ 11200.19} &  \multicolumn{2}{c|}{12552.96 } & \multicolumn{2}{c|}{12838.95 } \\
\hline
\multirow{2}*{S3} & \multicolumn{2}{c|}{ --}&Source &Hybrid &Source &Hybrid  \\
\cline{2-7}
& \multicolumn{2}{c|}{ --}&12055.02 &1860.34 & 12592.76 &985.92 \\
\hline
\end{tabular}}
}
\end{table}

Several other interesting phenomena can be observed by comparing Table \ref{tab:time} and Table \ref{tab:time2}. We can observe that when the feature distributions of the source data and target data are similar, the accuracy of the DDNN-based VAD drops slightly with respect to the depth of the network. One possible explanation is that the distributions can be sufficiently covered by the source corpus, so that we can achieve a desired performance with just one hidden layer of DDNN. On the contrary, when the feature distributions are dissimilar, the accuracy increases dramatically with respect to the depth of the network, which demonstrates the power of the transfer learning schemes.
However, when compared with the referenced methods, we can observe that when the source and target environments are relatively similar, the DDNN-based VAD outperforms the referenced methods. But when the environments are severely dissimilar, the DDNN-based VAD is weaker than the referenced ones.

Table \ref{tab:time3} lists the pre-training time comparison between the schemes. From the table, we can see that Scheme 1 is the most efficient one, and Scheme 3 is slightly slower than Scheme 1.

\section{Conclusions}
\label{sec:conclusion}
In this paper, we have tried to solve the mismatching problem between the source corpus and the target corpus in the transfer learning perspective, and further tried three DDNN-based domain adaptation schemes for the problem. Experimental results have shown that Schemes 1 and 2 are effective in dealing with the mismatching problem of VAD when compared with the traditional training method, while Scheme 1 is much more efficient than Scheme 2. The results also have shown that the layer-wise pre-training strategy is important for the success of the deep-learning-based transfer learning schemes. Although Scheme 3 is failed, it does provide an attempt on the compromise between the training time and accuracy, and provide a contrary example for showing the effectiveness of the layer-wise pre-training.

Experimental results also have shown that when the source and target corpora are very dissimilar, the performance might be weaker than the referenced methods. For solving this, pre-training with more unlabeled target data, with multiple source domain, and with more hidden layers might be helpful. Moreover, how to make the performance of Scheme 2 more closer to the upper bound, how to accelerate Scheme 2 and meanwhile keep its effectiveness are also what we want to address. We leave these problems as the future work.


%
\eightpt
\bibliographystyle{IEEEtran}
\bibliography{zxlrefs}
\end{document}